\documentclass{article}

\usepackage{microtype}
\usepackage{graphicx}
\usepackage{subfigure}
\usepackage{booktabs} 

\usepackage{hyperref}

\usepackage[accepted]{icml2024}

\usepackage{amsmath}
\usepackage{amssymb}
\usepackage{mathtools}
\usepackage{amsthm}

\usepackage{multirow}

\usepackage[capitalize,noabbrev]{cleveref}

\theoremstyle{plain}

\theoremstyle{definition}

\theoremstyle{remark}

\usepackage[textsize=tiny]{todonotes}

\icmltitlerunning{SuperPose: Improved 6D Pose Estimation with Robust Tracking and Mask-Free Initialization}

\begin{document}

\twocolumn[
\icmltitle{SuperPose: Improved 6D Pose Estimation with Robust Tracking and Mask-Free Initialization}



\icmlsetsymbol{equal}{*}

\begin{icmlauthorlist}
\icmlauthor{Yu Deng}{yyy}
\icmlauthor{Jiahong Xue}{yyy}
\icmlauthor{Teng Cao}{yyy}
\icmlauthor{Yingxin Zhang}{zzz}
\icmlauthor{Lanxi Wen}{zzz}
\icmlauthor{Yiyang Chen}{zzz}
\end{icmlauthorlist}


\icmlkeywords{6D Pose Estimation, Industrial Automation, SAM2, LightGlue, Object Tracking}

\vskip 0.3in
]



\printAffiliationsAndNotice{\icmlEqualContribution} 

\begin{abstract}
    We developed a robust solution for real-time 6D object detection in industrial applications by integrating FoundationPose, SAM2, and LightGlue, eliminating the need for retraining. Our approach addresses two key challenges: the requirement for an initial object mask in the first frame in FoundationPose and issues with tracking loss and automatic rotation for symmetric objects. 
    The algorithm requires only a CAD model of the target object, with the user clicking on its location in the live feed during the initial setup. Once set, the algorithm automatically saves a reference image of the object and, in subsequent runs, employs LightGlue for feature matching between the object and the real-time scene, providing an initial prompt for detection. Tested on the YCB dataset and industrial components such as bleach cleanser and gears, the algorithm demonstrated reliable 6D detection and tracking. 
    By integrating SAM2 and FoundationPose, we effectively mitigated common limitations such as the problem of tracking loss, ensuring continuous and accurate tracking under challenging conditions like occlusion or rapid movement.
\end{abstract}

\section{Introduction}
Estimating the rigid 6D transformation between an object and the camera, also referred to as object pose estimation, is a critical task in various domains such as robotic manipulation \cite{kappler2018real, wen2022catgrasp, wen2022you} and mixed reality \cite{marchand2015pose}. Traditional methods \cite{he2020pvn3d, he2021ffb6d, labbe2020cosypose, park2019pix2pose, wen2020robust} are typically instance-level, meaning they only function with specific object instances defined during training. 
These instance-level approaches usually require a textured CAD model for generating training data and are not applicable to novel, unseen objects at test time. In contrast, category-level methods \cite{chen2020learning, lee2023tta, tian2020shape, wang2019normalized, zhang2022ssp} do not require CAD models but are constrained to objects within predefined categories from the training phase.

To overcome the limitations of both approaches, FoundationPose \cite{wen2024foundationpose} offers a hybrid solution capable of robust 6D pose estimation using both model-based and model-free techniques. However, FoundationPose requires a manually annotated mask of the object in the first frame, a process that is tedious and impractical for automated pipelines. Automating the mask annotation step or minimizing user interaction would significantly enhance its applicability in real-time settings. 
To address this limitation, we incorporated a segmentation algorithm, enabling users to perform live, image, and video-based pose estimation by providing only a CAD model. Additionally, this segmentation approach resolves issues related to tracking loss.

Our aim is to streamline the entire process, requiring users to provide only a CAD model. As such, the input to the segmentation process must also be derived from the CAD model. We conducted extensive experiments on CNOS \cite{nguyen2023cnos} and PerSAM \cite{zhang2023personalize}, but both algorithms exhibited shortcomings. While CNOS meets the CAD model requirement, its performance lacks consistency, particularly when the CAD model is imprecise, leading to significant loss of edge information in the final segmentation. 
PerSAM, on the other hand, demands a reference image and mask file, making it highly sensitive to variations in angle and occlusions, as it relies on cosine similarity to compare the reference and test images.

Ultimately, we found that SAM2 \cite{ravi2024sam} provided superior results. By simply clicking on the location of the object in the test image, the algorithm automatically generates a segmentation and a segmented image based on the user-specified object. For future runs, only the CAD model is needed, and the system leverages LightGlue \cite{lindenberger2023lightglue} to match feature points between the segmented image and the first frame of the live video or image. 
These matched points serve as location prompts for SAM2, enabling fully automated object tracking and segmentation without any additional user input. This approach not only addresses the mask annotation challenge but also facilitates a fully automated workflow for object pose estimation.

\begin{figure*}[htbp]
    \centering
    \includegraphics[width=\linewidth]{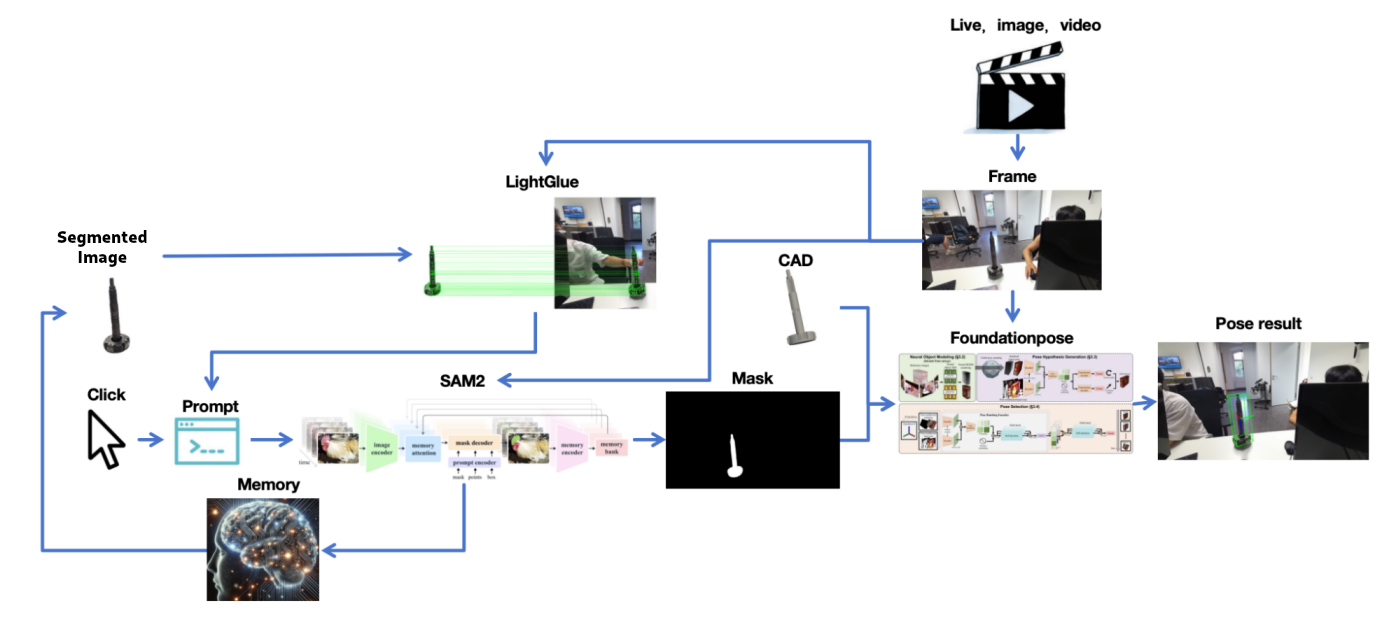}
    \caption{\textbf{System workflow} The image shows the implementation process of the entire system. In our system, the object's positional information within the real image is initially obtained either through user clicks or by employing LightGlue to perform feature matching between the segmented image and the real image, thus providing a positional prompt. 
    This information is then transmitted to SAM2. If a segmented image of the object does not exist, one is generated and stored in memory. Simultaneously, a mask matrix for the object is created. Subsequently, by integrating the generated mask, the object's CAD model, and the real frame, FoundationPose is ultimately utilized to perform 6D pose estimation.}
    \label{fig:workflow}
\end{figure*}

As shown in Figure 1, we only need to provide a CAD model and click on the target we want to recognize during the first use or provide a segmented image of the target. The system can then automatically complete the entire 6D pose estimation without any other operations.

We also tackled the problem of tracking loss in FoundationPose by incorporating a method that calculates the distance between centroids during 6D pose detection and segmentation. 
SAM2’s introduction of a memory module significantly enhances tracking performance, as demonstrated in our tests. Even if an object temporarily disappears from the scene and reappears, SAM2 is capable of effectively re-establishing the track. 
Despite this, SAM2 exhibits limitations when the object remains absent for an extended period, which can hinder its ability to regain detection and tracking. To mitigate this limitation, we employed feature point matching, allowing the system to re-identify the object and continue tracking reliably after prolonged absences.

Our main contributions are summarized as follows:

\begin{itemize}
    \item \textbf{A novel 6D object pose estimation system without manual mask annotation:} We introduce a system that requires only a single click and a CAD model for initialization, eliminating the need for labor-intensive mask annotations.
    \item \textbf{Live and stable pose estimation from images and videos:} Our approach enables efficient, live and stable estimation of object poses from both image and video data, enhancing applicability in dynamic environments.
    \item \textbf{Addressing key challenges for continuous and accurate tracking:} The system tackles critical issues such as pose re-registration and automatic orientation correction for symmetric objects, ensuring continuous and precise tracking over time.
\end{itemize}

\section{Related Work}
\subsection{CAD Model-Based Object Pose Estimation}
The first category of methods is instance-level pose estimation approaches \cite{he2020pvn3d, he2021ffb6d, labbe2020cosypose, park2019pix2pose}, which assume that a textured CAD model of the object is available. These methods are trained and tested on the exact same object instance. Pose estimation can be performed through direct regression \cite{li2019cdpn, xiang2017posecnn} or by leveraging Perspective-n-Point (PnP) algorithms \cite{park2019pix2pose, tremblay2018deep}, which construct 2D-3D correspondences. Additionally, 3D-3D correspondences can be used to solve the object pose through least-squares fitting techniques \cite{he2020pvn3d, he2021ffb6d}.

In contrast, category-level methods \cite{chen2020learning, lee2023tta, tian2020shape, wang2019normalized, zhang2022ssp} enable pose estimation for novel objects within the same category. However, these methods are limited to predefined categories and cannot generalize to arbitrary objects outside of these categories.

A third approach \cite{labbe2022megapose, shugurov2022osop} aims to estimate the pose of non-predefined objects by using only a CAD model and a mask of the object in the first frame. FoundationPose is a representative example of this category. It uses a synthetic dataset generated with the help of large language models (LLMs) and employs a Transformer-based network architecture combined with contrastive learning, achieving state-of-the-art results on datasets such as YCB. 
However, FoundationPose\cite{wen2024foundationpose} does not provide a built-in tool for mask generation, often requiring manual annotation. Additionally, it lacks the capability for real-time pose estimation, and tracking loss can occur when the object moves rapidly or becomes occluded.

\subsection{Instance Segmentation}
Most instance-level segmentation algorithms \cite{he2017mask, yurtkulu2019semantic} require fine-tuning when applied to specific datasets, such as gears or objects outside predefined categories. This process can result in significant redundancy and demand substantial human and computational resources. 
Typically, it involves collecting a new dataset tailored to the segmentation task, annotating it, and retraining the model while monitoring for convergence.

Recently, with the introduction of the Segment Anything Model (SAM) \cite{kirillov2023segment}, several train-free algorithms have emerged, including PerSAM \cite{zhang2023personalize} and CNOS \cite{nguyen2023cnos}. PerSAM utilizes cosine similarity between a reference image, processed through the SAM algorithm, and a test image to perform segmentation. 
CNOS, given a CAD file, generates images from various angles and uses them as inputs to the SAM algorithm for segmentation. However, both methods have notable limitations. PerSAM is highly sensitive to variations in the object’s angle and still requires additional data, such as a mask, for initialization. CNOS, while effective in theory, is particularly sensitive to the CAD file; if the texture and details of the CAD model significantly differ from the real object, the segmentation process may fail.

A newly introduced algorithm, SAM2 \cite{ravi2024sam}, offers improvements through its memory mechanism, allowing for instance segmentation and real-time tracking with minimal input, such as a prompt. SAM2 is capable of re-tracking an object even after temporary disappearance, provided it reappears within a short timeframe, making it a more robust solution for real-time applications.

\subsection{Feature Point Matching}
Traditional image matching algorithms rely on hand-crafted criteria and gradient statistics \cite{lowe2004distinctive, harris1988combined, bay2006surf, rosten2006machine}. In recent years, however, much research has shifted toward using Convolutional Neural Networks (CNNs) for feature detection \cite{yi2016lift, DeTone_2018_CVPR_Workshops, DBLP:journals/corr/abs-1905-03561, NEURIPS2019_3198dfd0, NEURIPS2020_a42a596f} and description \cite{NIPS2017_831caa1b, Tian_2019_CVPR}. 
CNN-based approaches have significantly enhanced the accuracy of feature matching. Some algorithms improve feature localization \cite{lowe2004distinctive}, while others offer high repeatability \cite{DeTone_2018_CVPR_Workshops}. Certain methods reduce storage and matching costs \cite{rublee2011orb}, some are invariant to specific transformations \cite{pautrat2020online}, and others ignore unreliable features \cite{NEURIPS2020_a42a596f}. These techniques typically rely on nearest neighbor search in descriptor space to match local features. 
However, non-matchable keypoints and imperfect descriptors can result in erroneous correspondences.

Deep matchers, such as SuperGlue \cite{sarlin2020superglue}, are trained neural networks designed to jointly match local descriptors and reject outliers based on an input image pair. SuperGlue, which combines Transformers \cite{vaswani2017attention} and optimal transport theory \cite{peyre2019computational}, leverages scene geometry and camera motion priors to achieve robust performance under extreme conditions. However, training SuperGlue is challenging due to its computational complexity, which scales quadratically with the number of keypoints. 
As a result, the original SuperGlue’s long runtime often necessitates reducing the size of the attention mechanism to improve efficiency, though this compromises performance.

Another approach is LoFTR \cite{sun2021loftr}, which matches points on dense grids rather than sparse locations. This method significantly improves robustness but comes at the cost of slower processing, as it must handle a larger number of elements, thereby limiting the resolution of the input image.

LightGlue, by contrast, surpasses existing methods such as SuperGlue in both speed and efficiency when matching sparse features. Its adaptive stopping mechanism allows for fine-grained control over the trade-off between speed and accuracy, making it possible to train high-performance deep matchers even with limited computational resources. LightGlue achieves Pareto optimality in balancing efficiency and accuracy, providing a versatile and effective solution for feature matching tasks.

\section{Proposed Method}

Our system integrates SAM2, LightGlue, and FoundationPose to achieve 6D pose estimation from a CAD model. The framework is illustrated in Figure 1, which outlines the entire process and the generated results. 
In the following subsections, we explain the system in detail, focusing on how it addresses key challenges.

\subsection{Integration and Testing of Methods}
We combine SAM2, LightGlue, and FoundationPose to achieve 6D pose estimation, instance segmentation, and feature point matching using only a CAD model file. The detailed implementation process is outlined as follows:

\begin{figure}
    \centering
    \includegraphics[width=\linewidth]{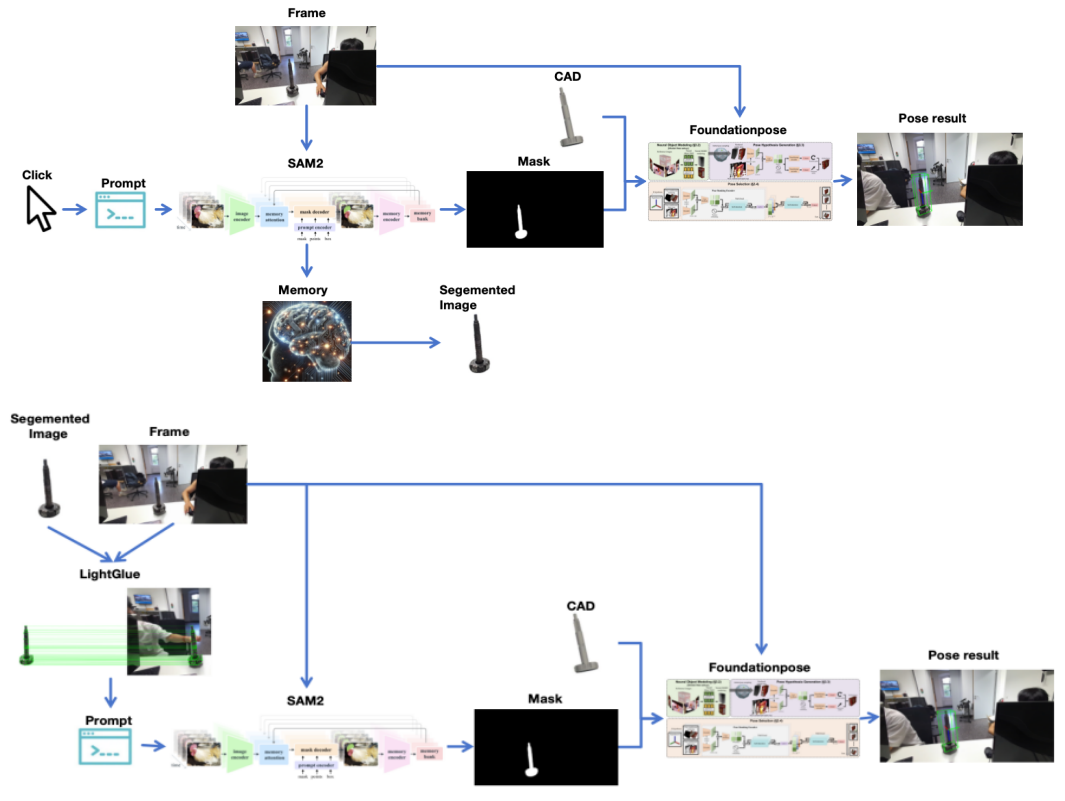}
    \caption{\textbf{The initialization via manual selection or segmented image input} The image illustrates the initialization process of our system. In the first method, users simply click on the frame, which sends a prompt to SAM2. This generates a mask that is passed to FoundationPose to produce a pose estimation and a segmented image, which is then stored for the memory mechanism. In the second method, users provide a segmented image to LightGlue, which generates a prompt for SAM2. SAM2 then produces a mask, which is used by FoundationPose to generate the estimation.}\label{initial}
\end{figure}

\subsubsection{Initial Target Identification via Manual Selection or segmented Image Input}
The system requires the user to provide a CAD model of the object to be recognized. During the first run, the user can either manually select the target object by clicking on it or provide a segmented image of the object as demonstrated in Figure \ref{initial}.
\subsubsection{Segmentation Prompt Generation Based on User Input or Feature Matching}
Once the target object is identified via manual selection or segmented image input(an image of the object with a pure white background), the system generates prompt information used by SAM2 for segmentation. 
SAM2, trained on the SA-V dataset\cite{ravi2024sam}, utilizes a Transformer-based encoder with ViT\cite{ravi2024sam} to extract image features and applies a Memory Attention mechanism to handle dependencies in video data. 
For mask, point, and box-type prompts, the encoder encodes these prompts, and the resulting representations, along with image features, are fed into a mask decoder to generate the mask. 
Thus, the object’s location, determined either by the user’s click or through feature point matching via LightGlue, serves as a point-type prompt for SAM2’s mask decoder, which then generates the final object mask.

\subsubsection{6D Pose Estimation}
Using the mask and CAD model, a rough pose estimation is initially performed through global sampling. The sampled poses and the cropped image are then fed into the Encoder and Conv ResBlock. 
The Encoder extracts the rotation and translation vectors. For pose hypothesis selection, images generated from different poses, along with the cropped region of the original image, are input into the Encoder to obtain feature representations, which are then pooled. 
Self-attention and multi-head attention mechanisms, followed by a fully connected layer, are used to score each pose hypothesis, with the highest-scoring hypothesis selected as the final pose estimate.

\subsection{Handling Tracking Loss in FoundationPose}
FoundationPose relies on positional information from the previous frame for pose estimation, which can result in tracking loss if the object moves too quickly. To address this, SAM2 performs real-time object segmentation for each frame. 
\begin{figure}
    \centering
    \includegraphics[width=\linewidth]{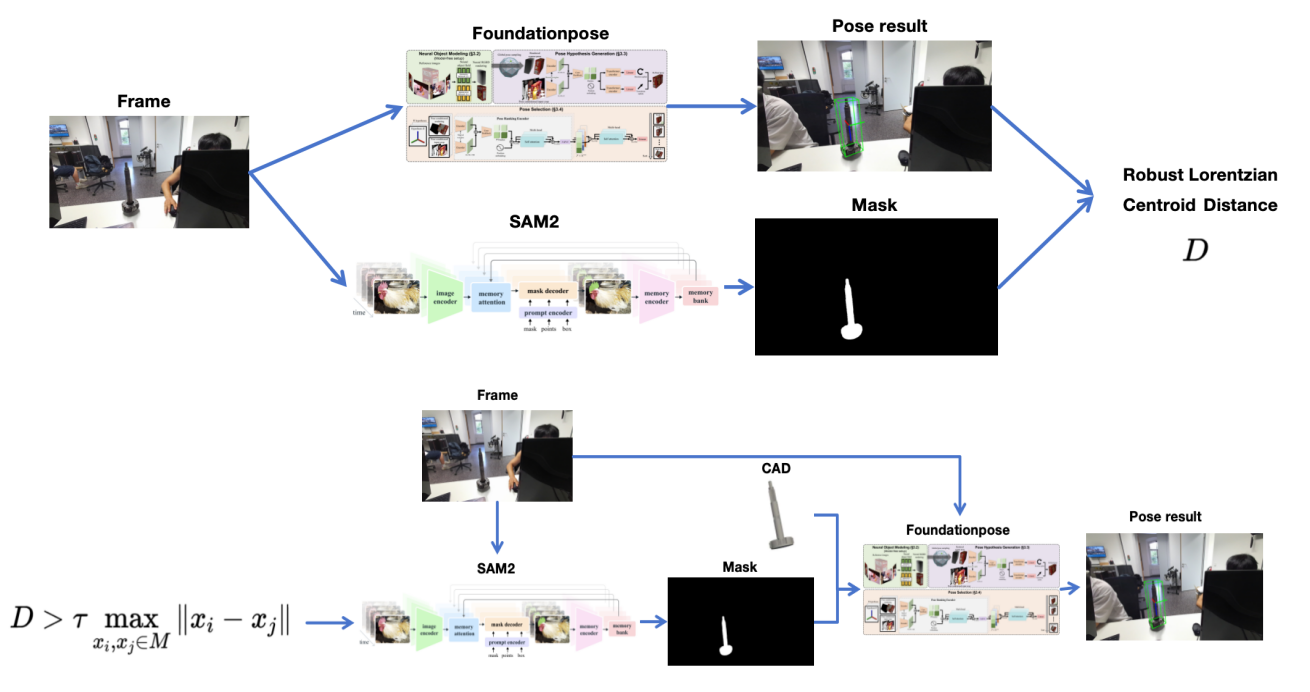}
    \caption{\textbf{The Process of Robust tracking} The image illustrates the process of robust tracking. In the first step, the frame is simultaneously sent to FoundationPose to obtain the pose estimation and to SAM2 to generate a mask. Afterward, the robust Lorentzian centroid distance is calculated. If this distance exceeds a predefined threshold, the system re-registers to solve the problem of tracking loss.}\label{robusttracking}
\end{figure}
We calculate the distance between the centroid of the mask and the center point of the pose estimate to detect tracking loss as demonstrated in Figure \ref{robusttracking}. In Eq.\ref{rld}, this distance is measured using Robust Lorentzian centroid Distance with Lorentzian error function and square of L2 norm.

\begin{equation}
    D = \log \left( 1 + \frac{1}{2 \sigma^2} \left\| \frac{1}{n} \sum_{i=1}^{n} \mathbf{r}_i - \mathbf{Hc} \right\|^2 \right) 
\label{rld}
\end{equation}

where  $\sigma$ is a parameter that regulates the sensitivity of the distance function to larger error values (outliers). The vector $\mathbf{r}_i$ corresponds to each point on the largest contour in the mask, while $\mathbf{c}$ represents the 3D center point obtained from the pose estimation. 
The matrix $\mathbf{H}$ is the transformation matrix that converts the 3D coordinate system to the image coordinate system.

As defined in Eq.\ref{exceed}, if the distance exceeds a certain threshold, the current mask generated by SAM2 is used to re-register the object, thereby correcting the tracking loss.

\begin{equation}
    D > \tau \max_{x_i, x_j \in M} \| x_i - x_j \| 
\label{exceed}
\end{equation}

where  $\tau$  is the coefficient of the maximum diameter, typically set to 0.2, and  $x_i$,  $x_j$  represent the vectors of points on the 3D model.

Additionally, we found that providing a higher-quality mask during registration mitigates automatic rotation and results in more stable tracking.

\subsection{Addressing Long-Term Object Loss: A Memory Mechanism}

\begin{figure}
    \centering
    \includegraphics[width=\linewidth]{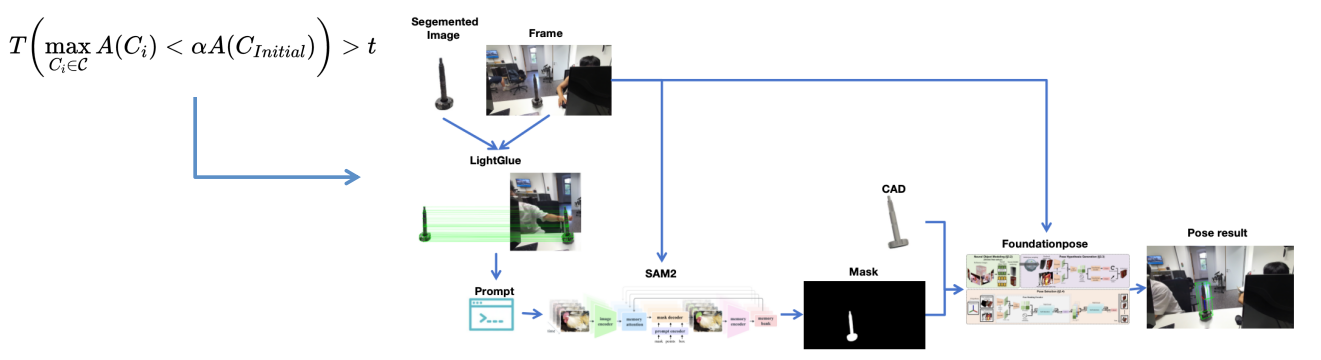}
    \caption{\textbf{The process of memory mechanism} The image depicts the memory mechanism process when an object is lost in the frame for an extended period. This is determined by measuring the difference between the area of the maximum contour and the initial contour. If the object remains lost for a specified duration, the segmented image is reloaded into LightGlue, which then passes a prompt to SAM2 to generate a mask for FoundationPose.}\label{memory}
\end{figure}

To handle long-term object loss, we implemented a memory-like mechanism demonstrated in Figure \ref{memory}. As defined in Eq.\ref{judgement}, if the object is lost for an extended period (typically 10 seconds), both FoundationPose and SAM2 may fail to track it.
\begin{equation}
    T \left( \max_{C_i \in \mathcal{C}} A(C_i) < \alpha A(C_{\text{Initial}}) \right) > t
\label{judgement}
\end{equation}

where  $C_i$  represents different contours in the mask image,  $A(C_i)$  is used to calculate the area of each contour, and  $C_{\text{Initial}}$  represents the largest contour in the mask during initialization.  
$\alpha$  is a parameter used to control the sensitivity of the memory mechanism, typically set to 0.6, and  $t$ is the time threshold, usually set to 10 seconds.

In such cases, we reload the segmented image of the object saved during the initial SAM2 run and perform feature point matching on the current frames using LightGlue. If the object reappears, a location prompt is generated, which is passed to SAM2 for re-segmentation. 
The segmentation result is then fed into FoundationPose for re-registration, allowing the system to resume tracking the object.


\section{Experiment}

\subsection{Deployment and Evaluation of FoundationPose}
To enhance object recognition and detection systems, we performed a series of experiments evaluating the performance of FoundationPose, focusing on its application to the YCB-Video dataset and industrial objects. The selected objects, such as detergent bottles (bleach cleanser) and gears, presented unique challenges due to their diverse shapes and textures.

FoundationPose was deployed in a live system using an Azure Kinect camera, which captured both RGB and depth data for real-time object detection and pose estimation in dynamic environments. However, several challenges arose during testing. 
First, FoundationPose requires accurate segmentation information, specifically a mask of the target object, to initiate the registration process. Generating this mask in real time proved difficult, highlighting the need for a real-time instance segmentation algorithm.

Furthermore, we observed tracking failures when objects moved too quickly or temporarily left the camera’s view, underscoring the need for more robust tracking mechanisms to maintain consistent detection. 
Another issue emerged when detecting gears; the system experienced rotation errors due to the symmetrical nature of the gears, resulting in incorrect orientation interpretations. This suggests that improved handling mechanisms for symmetrical objects are essential.

Addressing these challenges is critical to enhancing the system’s reliability and accuracy, particularly in complex, real-time operational environments.

\subsubsection{CNOS and CNOS+FoundationPose}
\begin{figure}
    \centering
    \includegraphics[width=\linewidth]{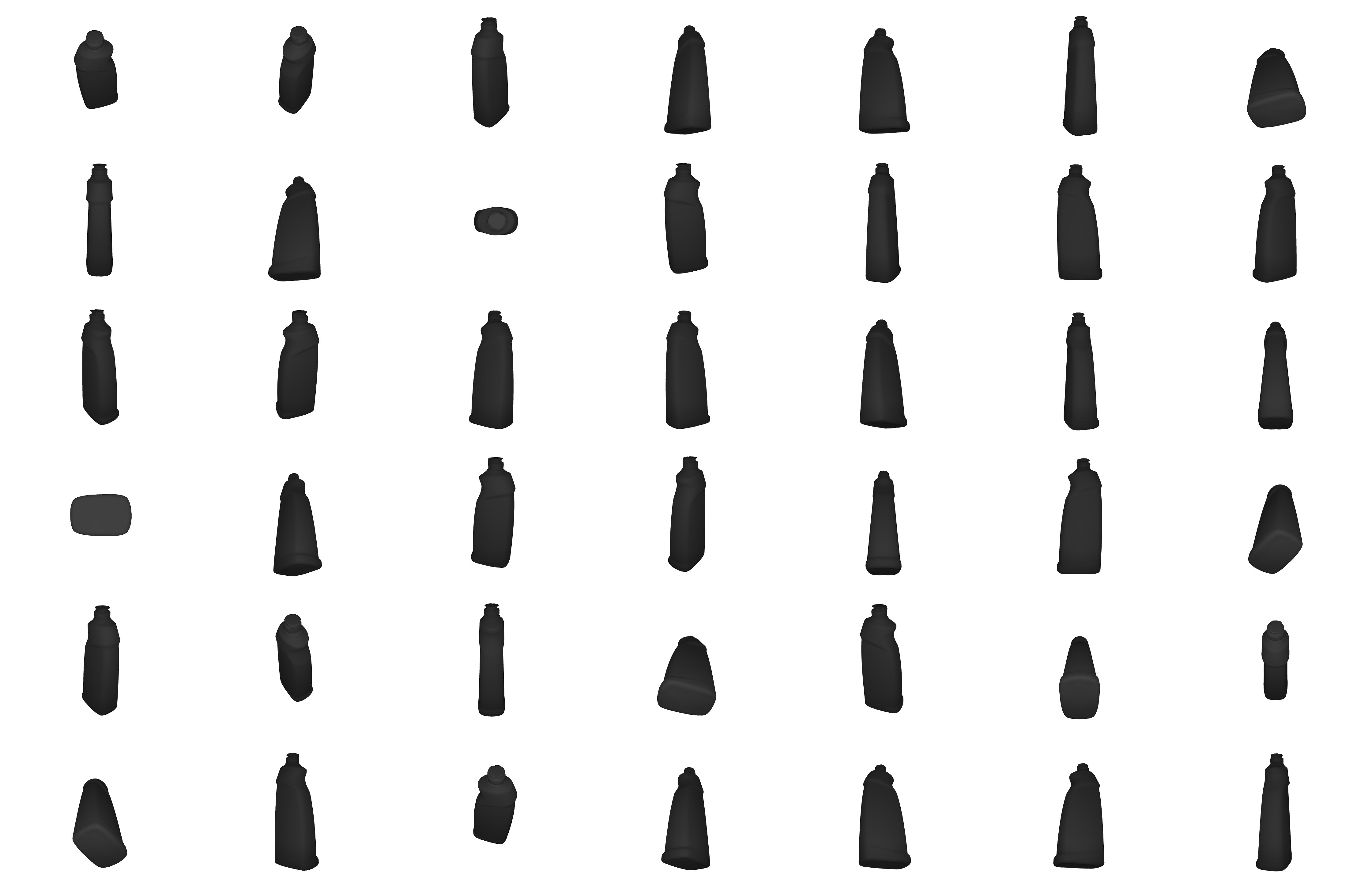}
    \caption{\textbf{The initial reference images generated from CNOS} \\The results demonstrate that CNOS initially generates 42 reference images from the CAD model as prompts, which are then passed to SAM.}\label{cnos_result}
\end{figure}

In our approach to instance segmentation, we initially selected CNOS due to its capability of performing segmentation using only the CAD model of the object, which is essential for tasks that rely on accurate geometric properties defined by the CAD model. 
We tested CNOS on various objects, including a bleach cleanser, and found that it effectively detected the object’s location and approximate shape. However, a significant limitation was identified with the edge representation. 
The segmented edges appeared jagged due to CNOS’s process of downsampling the input image to 224x224 pixels before upsampling it back to the original resolution, leading to a loss of fine edge details. Despite this limitation, CNOS remained functional within our system.

We integrated CNOS and FoundationPose into separate modules, with a workflow designed such that when the camera captures data and registration is required, CNOS generates a mask matrix, which is then passed to FoundationPose for registration and tracking. 
This pipeline delivered satisfactory results with fast processing speeds for the bleach cleanser, with the primary time cost occurring during initialization when CNOS rendered the CAD model to generate images from multiple angles.

However, a major challenge arose when testing CNOS on a gear object. It frequently failed to generate an accurate mask, which we traced to discrepancies between the CAD model and the real object, particularly regarding texture and material properties, despite the shape and edge information being consistent. 
These limitations, especially with complex objects like gears, highlighted that relying solely on CNOS would not be sufficient for all scenarios.

To overcome these limitations, we propose exploring alternative segmentation models and implementing a polymorphic approach within the mask class, allowing the system to switch between different models as needed. 
This modular design will enhance the system’s robustness and ensure high accuracy across a diverse range of objects and segmentation challenges.

\begin{figure}[H]
    \centering
    \begin{minipage}{0.2\textwidth}
        \centering
        \includegraphics[width=\linewidth]{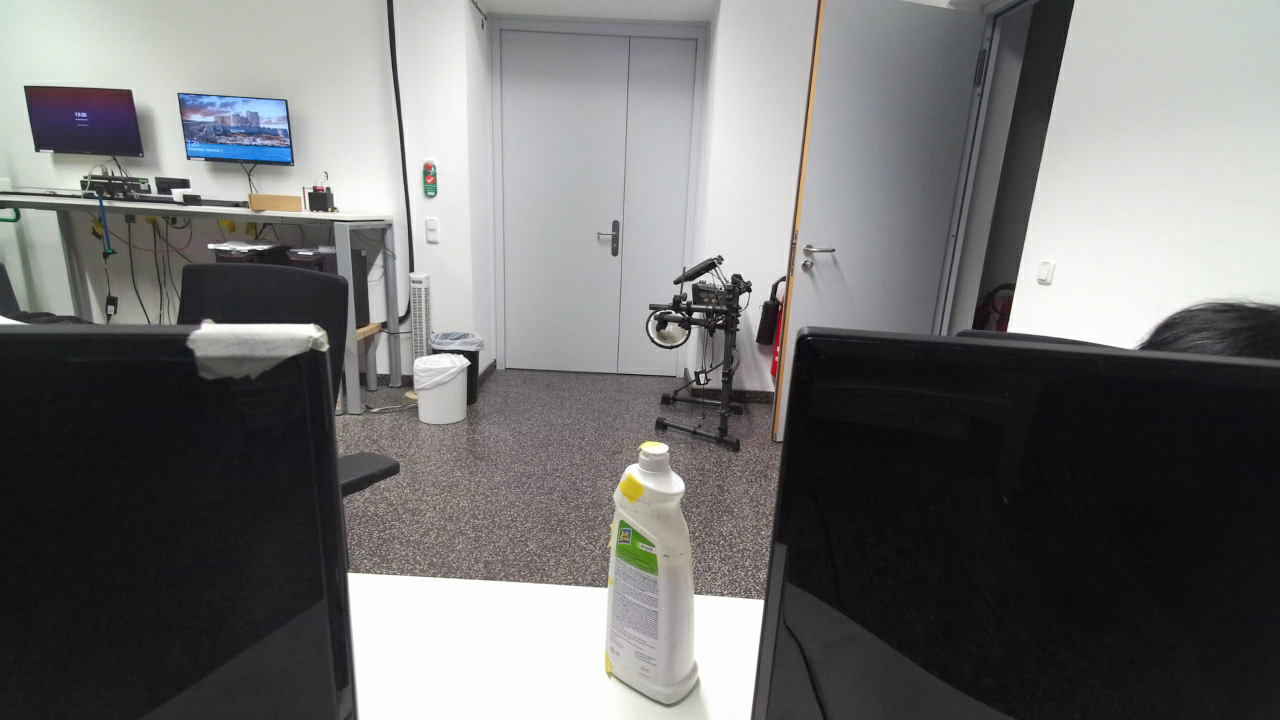}\label{fig:maskgear1}
    \end{minipage}
    \begin{minipage}{0.2\textwidth}
        \centering
        \includegraphics[width=\linewidth]{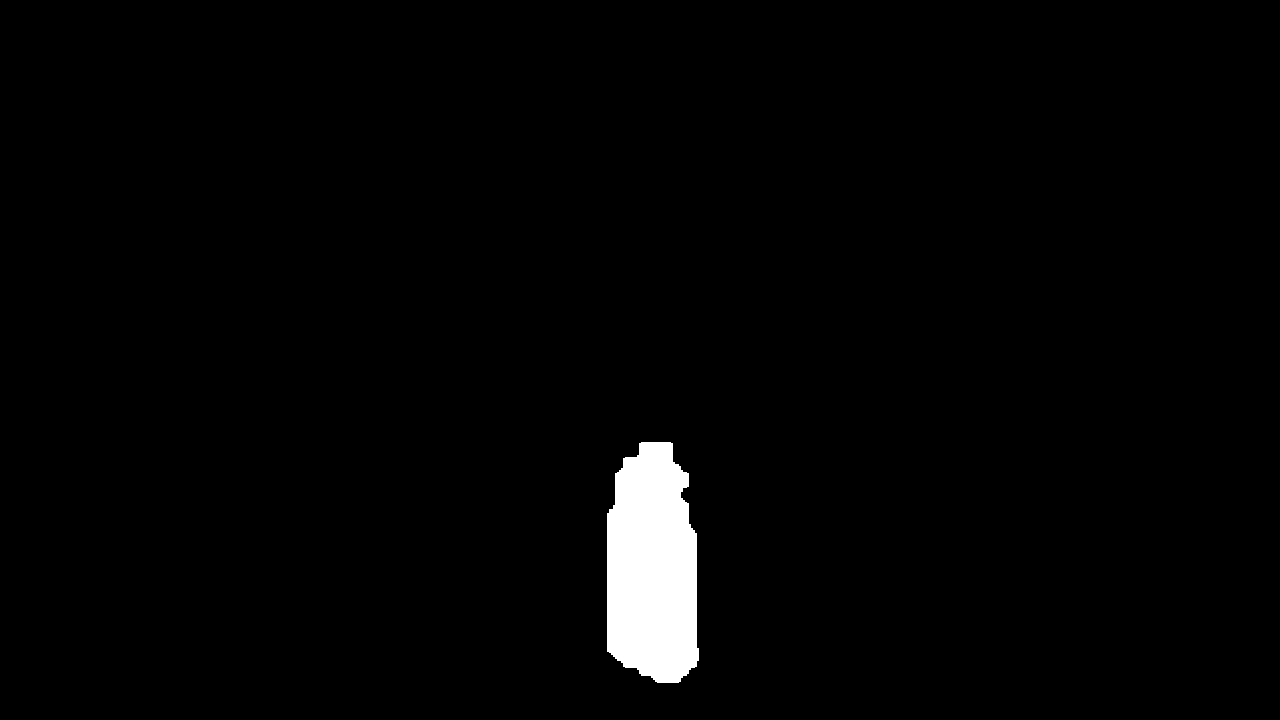}\label{fig:maskgear1}
    \end{minipage}
    \begin{minipage}{0.2\textwidth}
        \centering
        \includegraphics[width=\linewidth]{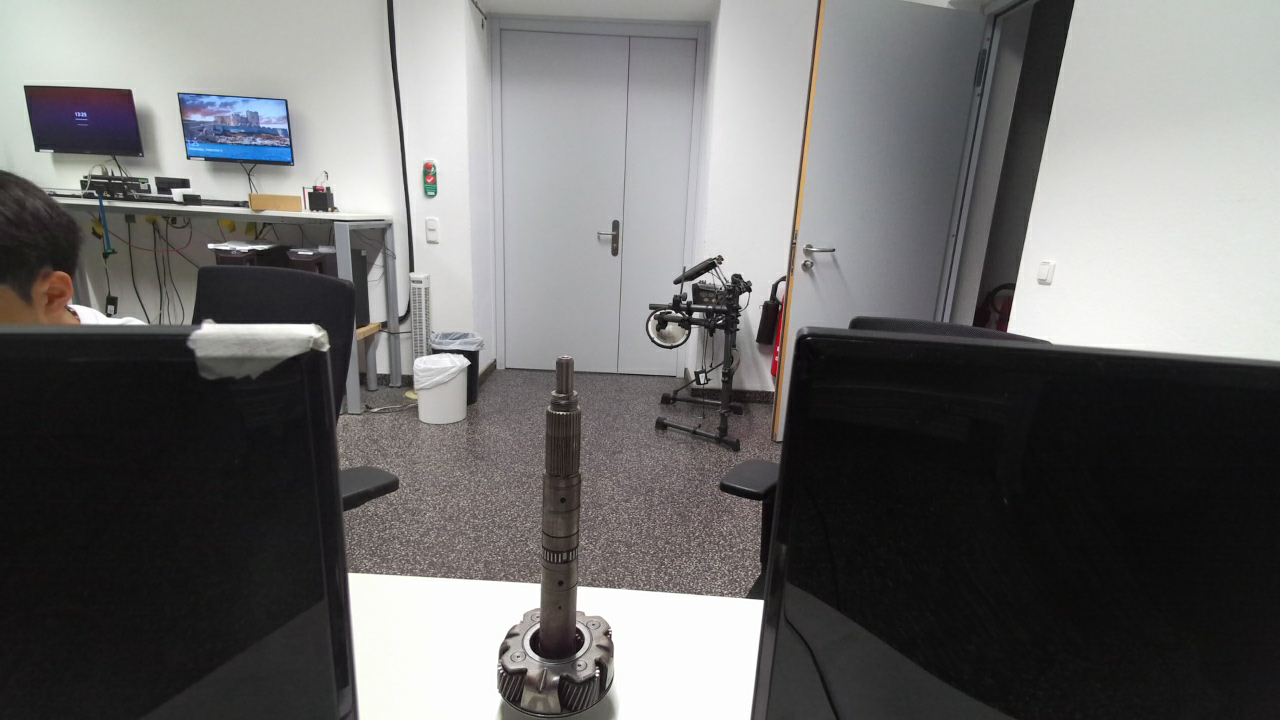}\label{fig:originalgear5}
    \end{minipage}
    \begin{minipage}{0.2\textwidth}
        \centering
        \includegraphics[width=\linewidth]{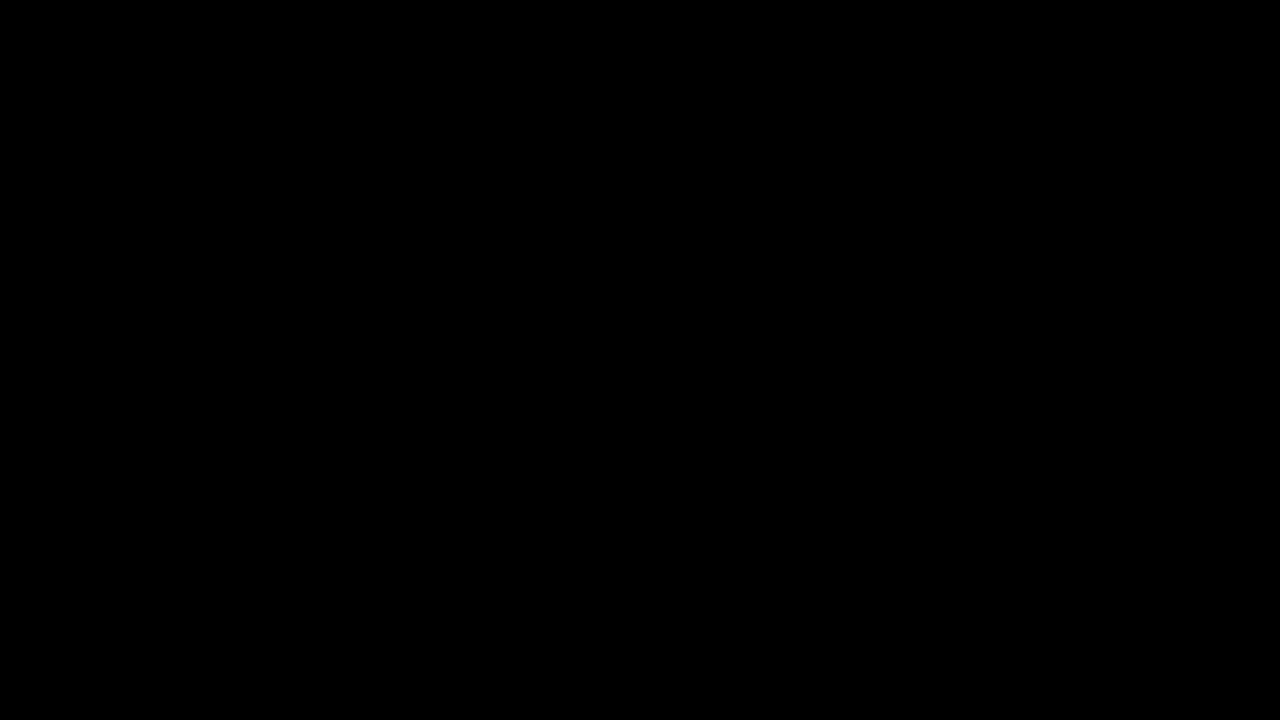}\label{fig:maskgear5}
    \end{minipage}
    
    \caption{\textbf{CNOS+FoundationPose Experimental Results} \\ As shown, CNOS successfully generates a mask for the bleach cleanser, though it tends to lose finer edge details. 
    In contrast, CNOS often fails to produce a valid mask for the gear, likely due to discrepancies between the CAD model and the real object’s texture and material properties.}
\end{figure}

\subsubsection{PerSAM and PerSAM+FoundationPose}
The Segment Anything Model (SAM) has recently gained significant attention in instance segmentation, and we explored the use of its variant, PerSAM, for generating mask information in our object detection system. 
PerSAM computes cosine similarity between a reference image and the current image to identify the target object’s mask. It demonstrated significant speed advantages over CNOS by eliminating the need for initialization, while offering more stable detections and effectively preserving edge details.

We integrated PerSAM into our system within the mask class, alongside FoundationPose. This integration improved the system’s overall robustness and accuracy across a variety of objects, including detergent bottles and gears. 
However, despite these improvements, several challenges emerged. First, PerSAM’s reliance on cosine similarity introduces instability due to its sensitivity to variations in viewing angles. This sensitivity can negatively affect tracking accuracy, particularly in dynamic environments where objects frequently change orientation.

Second, the requirement to provide a reference image and corresponding mask file for each object complicates the workflow, making it cumbersome, particularly in industrial settings where efficiency is critical. This added complexity can introduce delays, especially in fast-paced or large-scale operations.

While PerSAM offers superior speed and precision compared to CNOS, addressing the sensitivity to angle variations and simplifying the workflow are necessary steps to ensure consistent and efficient performance in real-world applications.

\begin{figure}[htbp]
    \centering
    \begin{minipage}{0.2\textwidth}
        \centering
        \includegraphics[width=\linewidth]{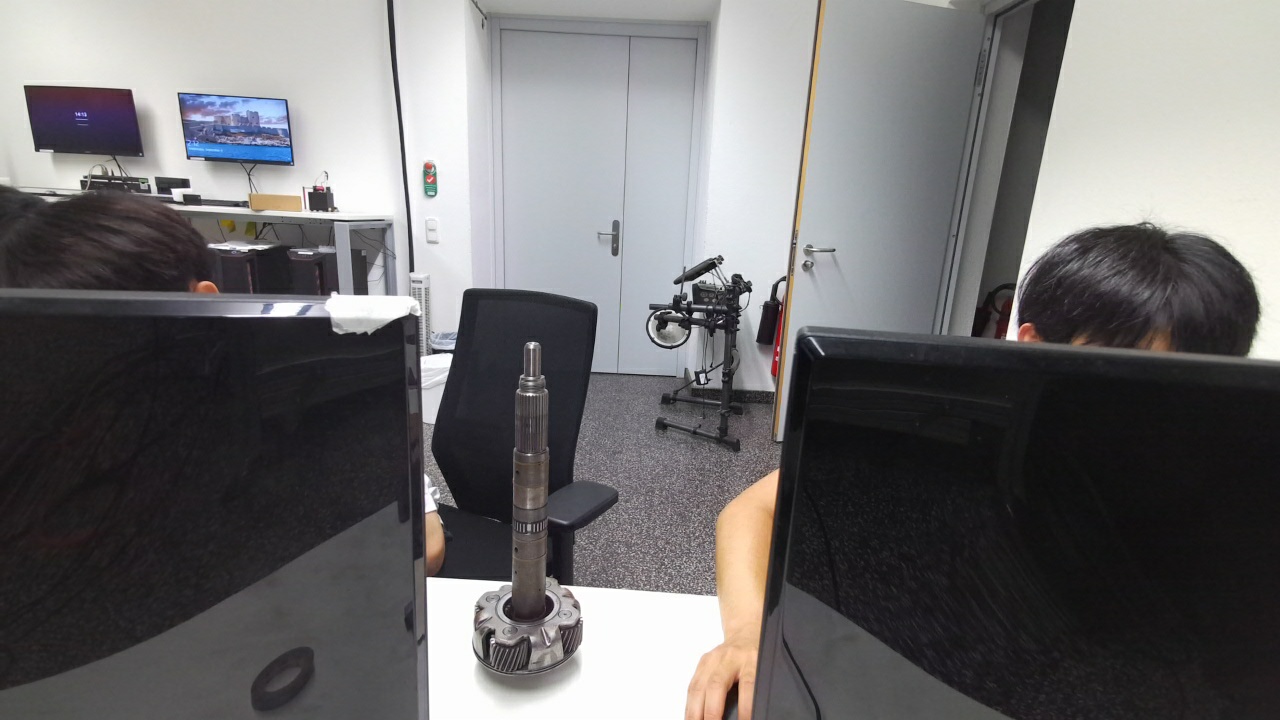}
        \label{fig:originalgear1}
    \end{minipage}
    \begin{minipage}{0.2\textwidth}
        \centering
        \includegraphics[width=\linewidth]{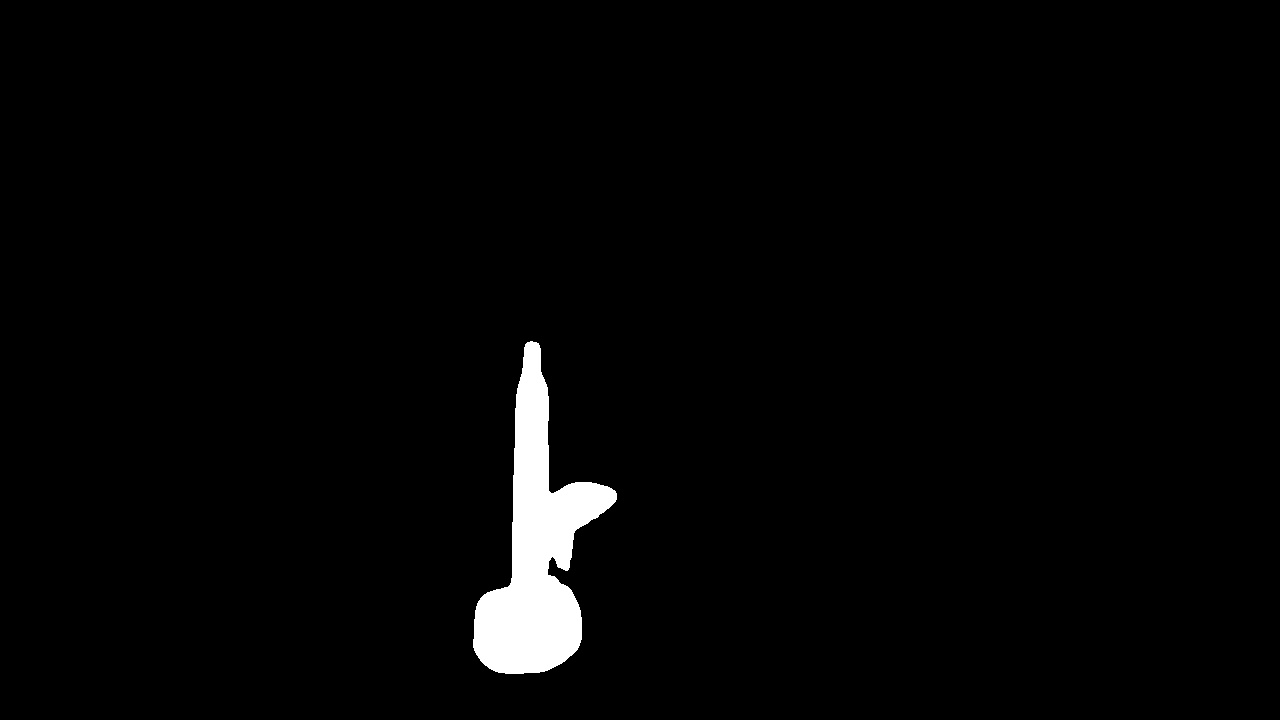}
        \label{fig:maskgear1}
    \end{minipage}
    \begin{minipage}{0.2\textwidth}
        \centering
        \includegraphics[width=\linewidth]{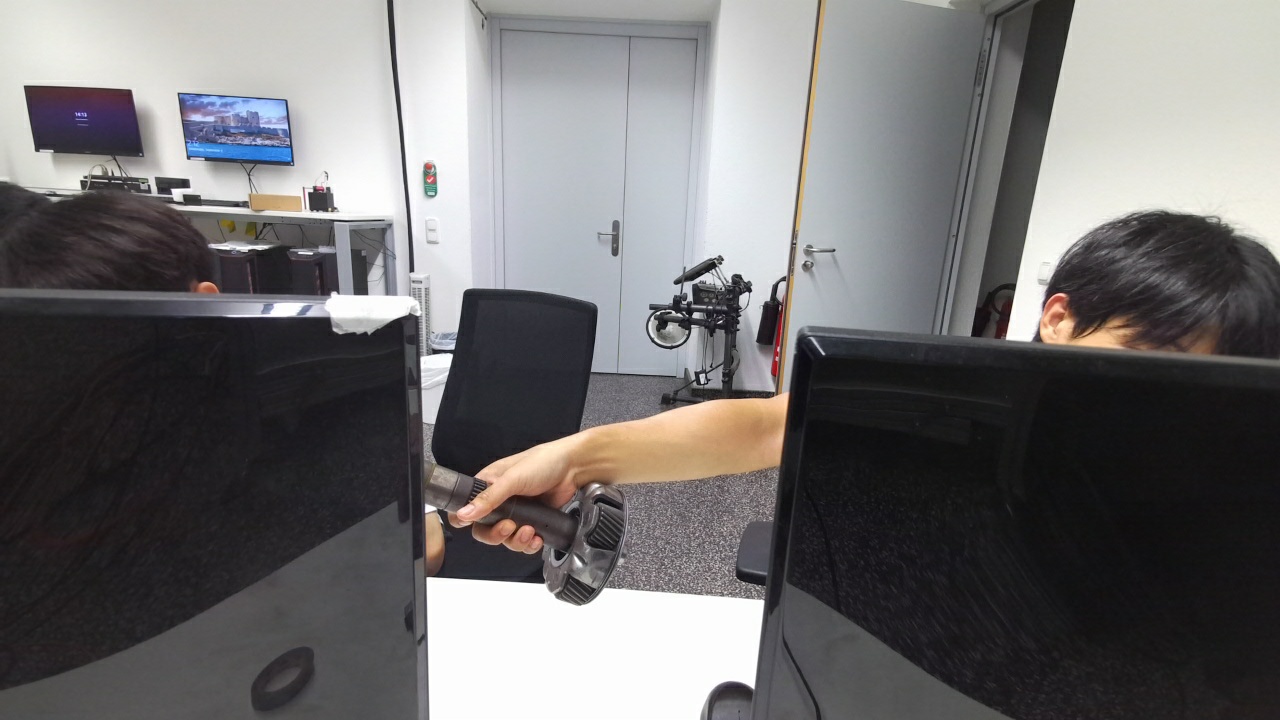}
        \label{fig:originalgear5}
    \end{minipage}
    \begin{minipage}{0.2\textwidth}
        \centering
        \includegraphics[width=\linewidth]{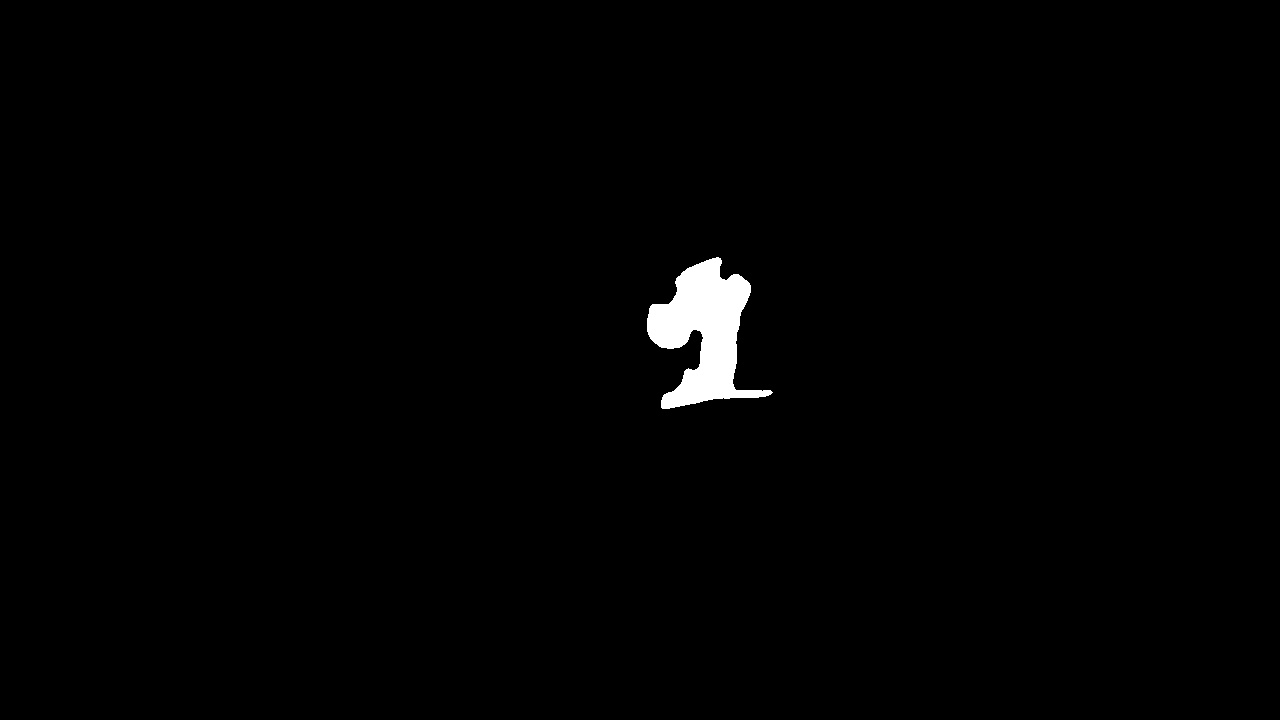}
        \label{fig:maskgear5}
    \end{minipage}
    
    \caption{\textbf{PerSAM+FoundationPose Experimental Results} \\PerSAM demonstrates superior performance in handling edge details, successfully recognizing both the gear and bleach cleanser in most cases. However, its sensitivity to variations in angles and occlusions often leads to misidentifications. 
    The figure above illustrates instances where edge details are lost and misidentifications occur.}
\end{figure}

\subsubsection{SAM2 and SAM2+FoundationPose}
To address the challenges identified in previous instance segmentation approaches, we adopted the recently introduced SAM2 algorithm, which provided significant improvements and effectively resolved key issues, including the loss of fine edge details, sensitivity to viewing angle variations, and the need for reference images and corresponding mask files. 
SAM2’s internal memory mechanism greatly enhances tracking capabilities, allowing it to maintain reliable object tracking even when objects temporarily disappear from the frame. Upon reappearance, SAM2 quickly and accurately reacquires the object, ensuring seamless tracking continuity—an improvement over earlier models that struggled with rapid orientation changes or when objects moved out of view.

Additionally, SAM2 simplifies the segmentation process with a click-to-segment operation, where the user selects the object at the beginning of a video, and SAM2 handles real-time segmentation and tracking throughout. This approach eliminates the need for pre-prepared reference images and masks, significantly reducing operational complexity.

Extensive experimentation demonstrated SAM2’s efficiency and accuracy in segmenting a variety of objects, including bleach cleansers and gears, with detection times averaging just 50 milliseconds per frame while delivering exceptional tracking accuracy. 
These advantages led to the integration of SAM2 into a dedicated class within the FoundationPose framework, resulting in a robust and efficient solution that overcomes the limitations of previous approaches. 

\begin{figure}[htbp]
    \centering
    \begin{minipage}{0.2\textwidth}
        \centering
        \includegraphics[width=\linewidth]{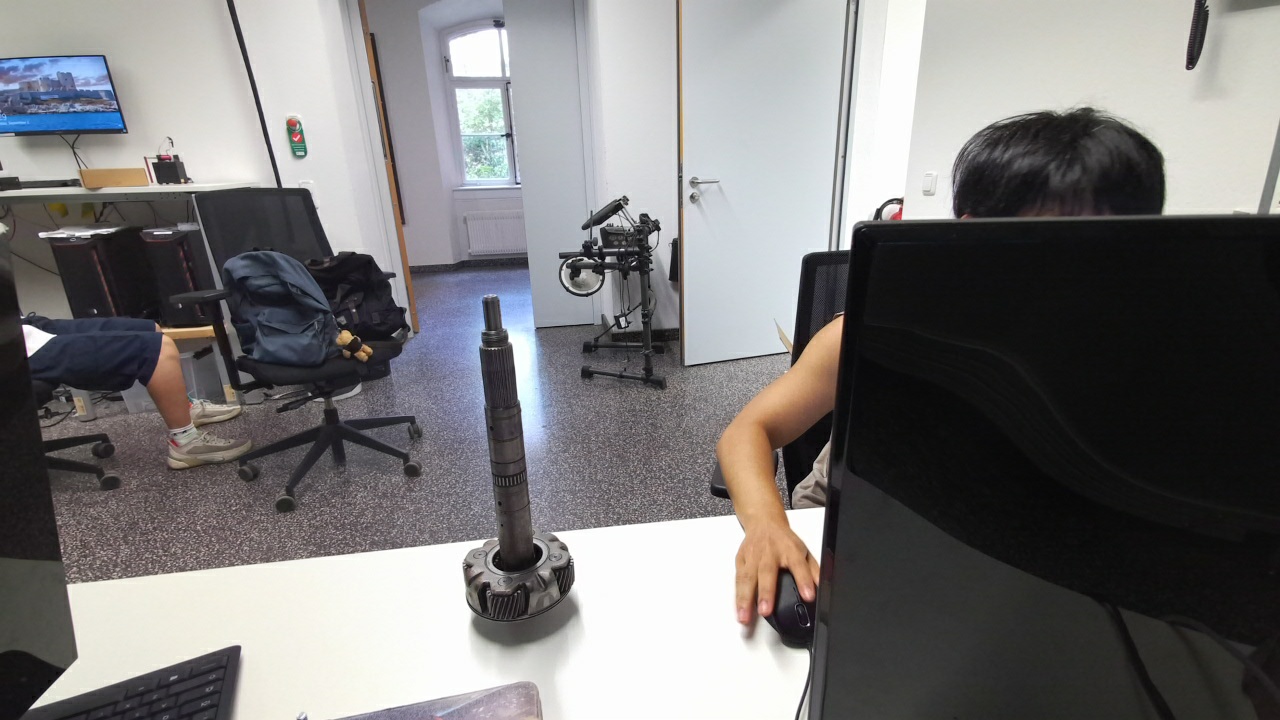}
        \label{fig:originalgear1}
    \end{minipage}
    \hfill
    \begin{minipage}{0.2\textwidth}
        \centering
        \includegraphics[width=\linewidth]{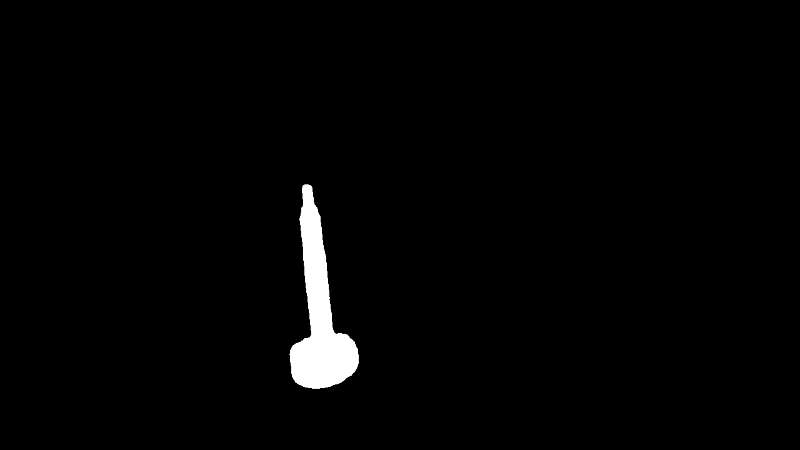}
        \label{fig:maskgear1}
    \end{minipage}
    \begin{minipage}{0.2\textwidth}
        \centering
        \includegraphics[width=\linewidth]{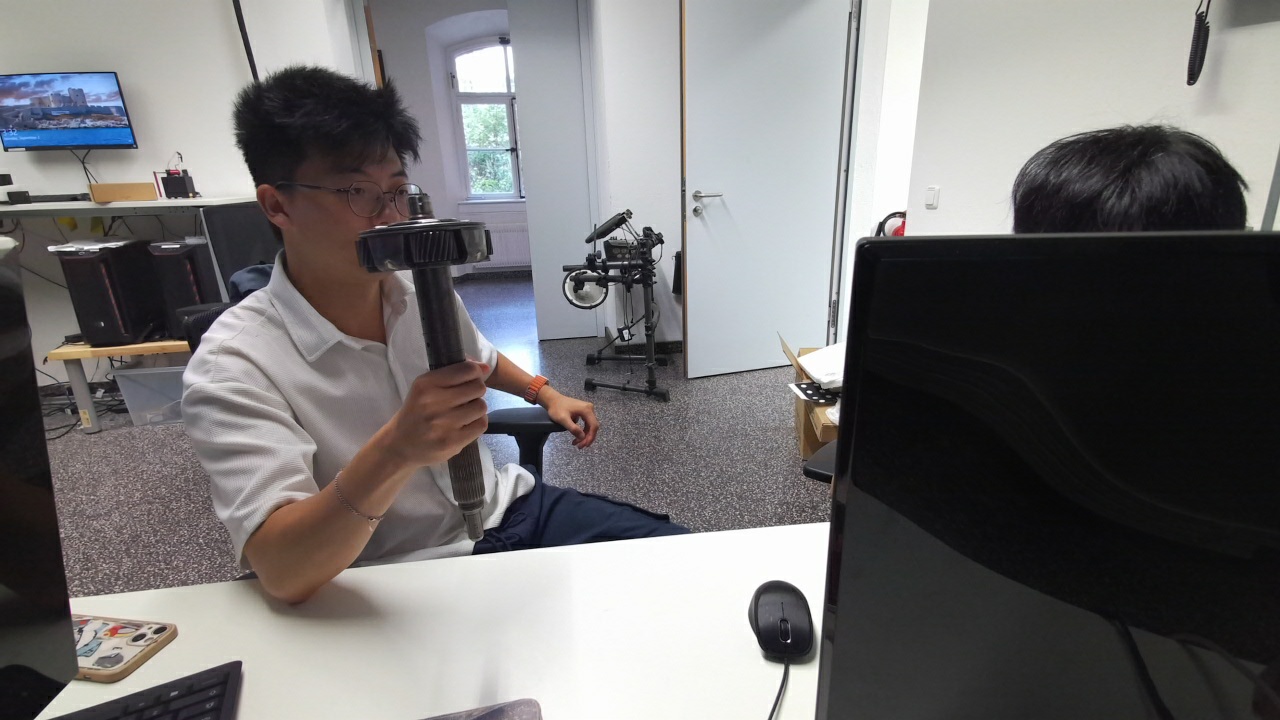}
        \label{fig:originalgear5}
    \end{minipage}
    \hfill
    \begin{minipage}{0.2\textwidth}
        \centering
        \includegraphics[width=\linewidth]{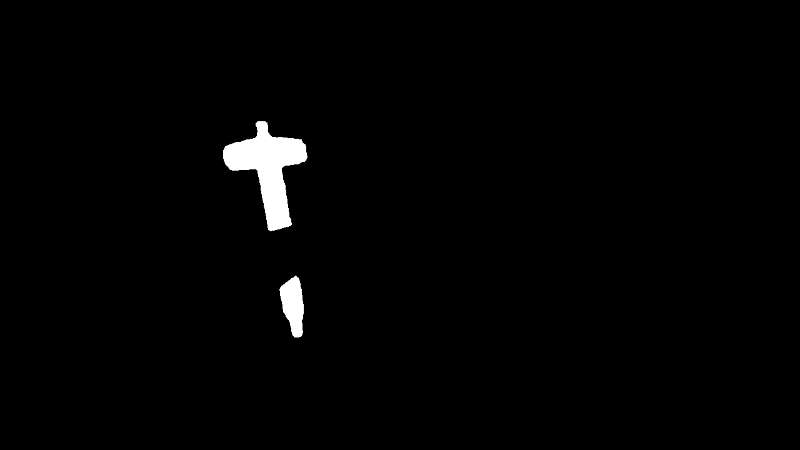}
        \label{fig:maskgear5}
    \end{minipage}
    
    \caption{\textbf{SAM2+FoundationPose Experimental Results} 
    \\
    The figure above illustrates the experimental results of SAM2 combined with FoundationPose. As shown, SAM2 excels in generating accurate masks for object segmentation, not only correctly identifying the target object but also preserving fine texture details.}
\end{figure}

SAM2 provides real-time segmentation and tracking with minimal user input, significantly enhancing the system’s practicality in dynamic environments. This integration represents a major advancement in object segmentation and tracking technology, ensuring the system is well-equipped for real-world deployment with high performance and reduced operational complexity.

\section{Experimental Result}

\subsection{Runtime of each model}
Runtime is a key metric used to evaluate the speed of various models. This analysis includes different segmentation models, such as CNOS, PerSAM, and SAM2, as well as different feature point matching models, SuperGlue and LightGlue. 
Additionally, the runtime of the base FoundationPose model, as well as FoundationPose enhanced with track loss resolving, is also considered.

\begin{table}[H]
    \setlength{\abovecaptionskip}{5pt}  
    \setlength{\belowcaptionskip}{0pt}  
    \resizebox{0.5\textwidth}{!}{  
    \begin{tabular}{|c|c|c|}
    \hline
    Algorithm & Initialised time(ms) & Track time for each frame(ms) \\ \hline
    CNOS & 5700 & 1200 \\ \hline
    SAM2 with tiny model & 250 & 48 \\ \hline
    PerSAM with FastSAM & 280 & 35 \\ \hline
    FoundationPose & 1200 & 100 \\ \hline
    LightGlue & 38 & 25 \\ \hline
    SuperGlue & 250 & 120 \\ \hline
    \end{tabular}
    }
    \caption{\textbf{Runtime of each model}
    }
    \label{tab:runtime_of_eahc_model}
\end{table}

Based on Table \ref{tab:runtime_of_eahc_model}, we compared the runtime performance of LightGlue versus SuperGlue, as well as CNOS, PerSAM, and SAM2.

LightGlue vs. SuperGlue: LightGlue has an initialization time of 38 milliseconds and a per-frame tracking time of 25 milliseconds, significantly faster than SuperGlue’s 250 milliseconds and 120 milliseconds. 
This indicates that LightGlue is more suitable for real-time applications.

Comparison of CNOS, PerSAM, and SAM2: CNOS has the highest initialization and per-frame tracking times, at 5700 milliseconds and 1200 milliseconds, respectively. 
PerSAM and SAM2 both have initialization times around 250 milliseconds, with per-frame tracking times of 35 milliseconds and 48 milliseconds. PerSAM slightly outperforms SAM2 in tracking speed.

LightGlue demonstrates superior runtime efficiency over SuperGlue, making it more appropriate for scenarios requiring high real-time performance. PerSAM and SAM2 have significantly lower runtimes compared to CNOS, with PerSAM being slightly faster in tracking. 
This suggests that when selecting an algorithm, one must balance performance and computational cost to meet specific requirements.

\subsection{IoU of each Segmentation model}
The Intersection over Union (IoU) metric evaluates the overlap between the predicted segmentation and the ground truth, which can be seen in below Eq.(\ref{AIoU}). 
We tested and compared the IoU results of SAM2, PerSAM, and CNOS on the YCB-Video dataset for segmentation tasks based on CAD models.

\begin{equation}
    \text{Average IoU} = \frac{1}{N} \sum_{i=1}^{N} \frac{|A_i \cap B_i|}{|A_i \cup B_i|}
\label{AIoU}
\end{equation}
where  $A_i$  represents the ground truth mask,  $B_i$  represents the predicted mask, and  $N$  represents the size of the dataset.

As shown in Table \ref{table:iou_values}, there are significant differences in the average IoU among the three algorithms: CNOS, PerSAM, and SAM2. 
Overall, SAM2 achieves the highest average IoU on most objects, demonstrating superior segmentation performance. For example, for the object \texttt{003\_cracker\_box}, SAM2 attains an average IoU of 0.8727, which is substantially higher than CNOS's 0.5409 and PerSAM's 0.3700. 
However, CNOS performs best on the object \texttt{009\_gelatin\_box}, achieving an average IoU of 0.9299, surpassing SAM2's 0.8900. This suggests that although SAM2 generally leads, there is still room for improvement on certain objects. PerSAM's average IoU is generally lower than the other two algorithms. 
In summary, while SAM2 exhibits the best overall performance, the algorithms display varying effectiveness across different objects.

\begin{table}[htbp]
    \centering 
    \resizebox{0.5\textwidth}{!}{ 
        \begin{tabular}{|l|c|c|c|}
        \hline
        \textbf{Object name} & \textbf{Average IoU (CNOS)} & \textbf{Average IoU (PerSAM)} & \textbf{Average IoU (SAM2)} \\ \hline
        002\_master\_chef\_can & 0.7567 & 0.4592 & 0.8501 \\ \hline
        003\_cracker\_box & 0.5409 & 0.3700 & 0.8727 \\ \hline
        004\_sugar\_box & 0.8232 & 0.5484 & 0.8525 \\ \hline
        005\_tomato\_soup\_can & 0.8222 & 0.5546 & 0.8523 \\ \hline
        006\_mustard\_bottle & 0.9097 & 0.7489 & 0.9184 \\ \hline
        007\_tuna\_fish\_can & 0.9007 & 0.4080 & 0.6771 \\ \hline
        008\_pudding\_box & 0.0406 & 0.3168 & 0.9299 \\ \hline
        009\_gelatin\_box & 0.9299 & 0.5126 & 0.8900 \\ \hline
        010\_potted\_meat\_can & 0.6965 & 0.3268 & 0.8060 \\ \hline
        011\_banana & 0.7003 & 0.7615 & 0.9110 \\ \hline
        019\_pitcher\_base & 0.8793 & 0.7160 & 0.8926 \\ \hline
        021\_bleach\_cleanser & 0.7395 & 0.7040 & 0.8473 \\ \hline
        024\_bowl & 0.4320 & 0.7117 & 0.8810 \\ \hline
        025\_mug & 0.7944 & 0.3696 & 0.8601 \\ \hline
        035\_power\_drill & 0.7590 & 0.3885 & 0.7910 \\ \hline
        036\_wood\_block & 0.6492 & 0.4904 & 0.8090 \\ \hline
        037\_scissors & 0.5551 & 0.2133 & 0.6892 \\ \hline
        040\_large\_marker & 0.7361 & 0.3070 & 0.7469 \\ \hline
        051\_large\_clamp & 0.6268 & 0.3632 & 0.8010 \\ \hline
        052\_extra\_large\_clamp & 0.4848 & 0.1611 & 0.7949 \\ \hline
        061\_foam\_brick & 0.4600 & 0.3396 & 0.8626 \\ \hline
        \textbf{MEAN} & \textbf{0.6779} & \textbf{0.4653} & \textbf{0.8350} \\ \hline
        \end{tabular}
    }
    \caption{\textbf{The comparison of IoU among different algorithms}}
    \label{table:iou_values}
\end{table}

\subsection{ADD of FoundationPose with different Segmentation model}
ADD (Average Distance of Model Points)\cite{xiang2017posecnn} is a metric commonly used to evaluate the accuracy of 6D object pose estimation. 
As shown in below Eq.\ref{ADD} it measures the average distance between corresponding 3D points on the ground truth model and the predicted model after transformation. 

\begin{equation}
    \text{ADD} = \frac{1}{m} \sum_{x \in M} \left\| (R x + t) - (R_{\text{gt}} x + t_{\text{gt}}) \right\|
\label{ADD}
\end{equation}

\begin{equation}
    \text{ADD-S} = \frac{1}{m} \sum_{x_1 \in M} \min_{x_2 \in M} \left\| (R x_1 + t) - (R_{\text{gt}} x_2 + t_{\text{gt}}) \right\|
\label{ADDS}
\end{equation}

where  $x$  represents the vectors of each point on the model,  $R$  represents the predicted rotation matrix,  $t$  represents the predicted translation vector,  $R_{\text{gt}}$  represents the ground truth rotation matrix,  $t_{\text{gt}}$  represents the ground truth translation vector, 
and  $m$  represents the total number of points in the model.
A pose is considered correct if the ADD is below a certain threshold.

\begin{equation}
    \text{Accuracy} = \frac{1}{N_{\text{total}}} \sum_{i=1}^{N_{\text{total}}} \mathbf{1}\left( \text{ADD}_i < \alpha \max_{x_j, x_k \in M} \|x_j - x_k\| \right)
\label{Accuracy}
\end{equation}

where  $\alpha$  represents the parameter for the maximum diameter, typically set to 0.1,  $x_j$  and  $x_k$  represent the vectors of points on the model, and  $N_{\text{total}}$  represents the size of the dataset.

However, for symmetric objects, the traditional ADD metric may not accurately measure pose estimation errors because symmetric objects can appear identical under certain rotations. 
To address this issue, the ADD-S (Average Distance of Model Points for Symmetric Objects) metric Eq.\ref{ADDS} is used. 

ADD-S calculates the average distance between each point on the predicted model and the closest point on the ground truth model, rather than between corresponding points. 
This approach more accurately evaluates the pose estimation accuracy for symmetric objects. A pose is considered correct if the ADD-S is below a specific threshold.

We tested the ADD accuracy Eq.\ref{Accuracy} on the YCB-Video dataset, comparing the results of FoundationPose with Ground Truth, FoundationPose with CNOS, FoundationPose with PerSAM, and FoundationPose with SAM2.

\begin{table}[htbp]
    \resizebox{0.5\textwidth}{!}{
    \begin{tabular}{|c|c|c|c|c|c|c|c|c|}
    \hline
    \multirow{2}{*}{\textbf{Algorithm}} & \multicolumn{2}{c|}{\textbf{FP+CNOS}} & \multicolumn{2}{c|}{\textbf{FP+PerSAM}} & \multicolumn{2}{c|}{\textbf{FP+Groundtruth}} & \multicolumn{2}{c|}{\textbf{FP+SAM2}} \\ \cline{2-9}
    \textbf{Metric} & \textbf{ADD-S} & \textbf{ADD} & \textbf{ADD-S} & \textbf{ADD} & \textbf{ADD-S} & \textbf{ADD} & \textbf{ADD-S} & \textbf{ADD} \\ \hline
    002\_master\_chef\_can & 95.67\% & 68.33 \% & 78.67\% & 52.33\% & 100.00\% & 70.33\% & 100.00\% & 70.33\% \\ \hline
    003\_cracker\_box & 66.22\% & 66.22\% & 52.00\% & 51.11\% & 100.00\% & 100.00\% & 100.00\% & 100.00\% \\ \hline
    004\_sugar\_box & 99.73\% & 99.73\% & 79.47\% & 78.93\% & 100.00\% & 100.00\% & 99.73\% & 99.73\% \\ \hline
    005\_tomato\_soup\_can & 77.78\% & 77.78\% & 67.56\% & 62.22\% & 95.11\% & 94.89\% & 95.56\% & 95.56\% \\ \hline
    006\_mustard\_bottle & 100.00\% & 97.33\% & 96.00\% & 94.67\% & 100.00\% & 98.00\% & 100.00\% & 98.00\% \\ \hline
    007\_tuna\_fish\_can & 99.67\% & 99.67\% & 54.00\% & 53.33\% & 100.00\% & 100.00\% & 99.67\% & 99.67\% \\ \hline
    008\_pudding\_box & 85.33\% & 85.33\% & 73.33\% & 73.33\% & 100.00\% & 100.00\% & 100.00\% & 100.00\% \\ \hline
    009\_gelatin\_box & 100.00\% & 100.00\% & 77.33\% & 77.33\% & 100.00\% & 100.00\% & 100.00\% & 100.00\% \\ \hline
    010\_potted\_meat\_can & 76.44\% & 76.44\% & 37.78\% & 29.33\% & 93.78\% & 80.89\% & 93.78\% & 80.89\% \\ \hline
    011\_banana & 77.33\% & 77.33\% & 84.67\% & 84.00\% & 100.00\% & 100.00\% & 100.00\% & 100.00\% \\ \hline
    019\_pitcher\_base & 98.67\% & 98.67\% & 90.67\% & 90.67\% & 100.00\% & 100.00\% & 100.00\% & 100.00\% \\ \hline
    021\_bleach\_cleanser & 83.33\% & 83.33\% & 87.00\% & 85.00\% & 100.00\% & 100.00\% & 100.00\% & 100.00\% \\ \hline
    024\_bowl & 50.00\% & 1.33\% & 94.67\% & 10.00\% & 100.00\% & 4.67\% & 100.00\% & 7.33\% \\ \hline
    025\_mug & 94.67\% & 92.00\% & 47.33\% & 42.67\% & 100.00\% & 98.00\% & 100.00\% & 98.00\% \\ \hline
    035\_power\_drill & 98.67\% & 98.67\% & 85.33\% & 84.67\% & 100.00\% & 100.00\% & 100.00\% & 100.00\% \\ \hline
    036\_wood\_block & 78.67\% & 8.00\% & 65.33\% & 6.67\% & 100.00\% & 8.00\% & 100.00\% & 9.33\% \\ \hline
    037\_scissors & 85.33\% & 85.33\% & 81.33\% & 80.00\% & 100.00\% & 100.00\% & 100.00\% & 100.00\% \\ \hline
    040\_large\_marker & 96.67\% & 54.00\% & 52.67\% & 19.33\% & 100.00\% & 52.67\% & 100.00\% & 52.00\% \\ \hline
    051\_large\_clamp & 78.00\% & 28.00\% & 54.67\% & 26.67\% & 100.00\% & 49.33\% & 100.00\% & 50.67\% \\ \hline
    052\_extra\_large\_clamp & 62.00\% & 8.67\% & 32.67\% & 4.00\% & 100.00\% & 17.33\% & 100.00\% & 18.00\% \\ \hline
    061\_foam\_brick & 61.33\% & 52.00\% & 41.33\% & 36.00\% & 100.00\% & 85.33\% & 100.00\% & 85.33\% \\ \hline
    \textbf{MEAN} & \textbf{85.94\%} & \textbf{75.22\%} & \textbf{69.60\%} & \textbf{58.52\%} & \textbf{99.13\%} & \textbf{84.12\%} & \textbf{99.13\%} & \textbf{84.32\%} \\ \hline
    \end{tabular}
    }
    \caption{\textbf{Pose tracking results of RGBD methods measured by AUC of ADD and ADD-S on YCB-Video dataset}\\
    }\label{table:pose_estimation}
\end{table}

As shown in the Table~\ref{table:pose_estimation}, significant differences exist in the ADD and ADD-S metrics among the three algorithms: FP+CNOS, FP+PerSAM, and FoundationPose+SAM2, across different objects. 
Overall, FoundationPose+SAM2 achieves the highest average ADD-S and ADD values on most objects, exhibiting superior performance in 6D pose estimation. For example, for the object \texttt{003\_cracker\_box}, 
FoundationPose+SAM2 attains 100.00\% in ADD-S and ADD, significantly surpassing FP+CNOS's 66.22\% and FP+PerSAM's 52.00\%.

FP+CNOS performs better on average than FP+PerSAM, with an average ADD-S of 85.94\% compared to FP+PerSAM's 69.60\%. However, FP+CNOS still shows shortcomings on certain objects such as object \texttt{051\_large\_clamp}.

Notably, the performance of FoundationPose+SAM2 is very close to that of FP+Groundtruth, with both achieving an average ADD-S of 99.13\% and average ADDs of 84.32\% and 84.12\%, respectively. 
This indicates that SAM2 can assist in 6D pose estimation to achieve results comparable to those obtained using ground-truth segmentation.

In summary, FoundationPose+SAM2 exhibits the best performance in the 6D pose estimation task, followed by FP+CNOS, with FP+PerSAM showing relatively lower performance.

\section{Conclusion}
In conclusion, we have successfully developed and implemented an algorithm that integrates FoundationPose, SAM2, and LightGlue, offering a robust solution for real-time and video-based 6D pose estimation. 
Our system represents a significant improvement over previous methods by eliminating the need for a pre-existing mask, resolving tracking loss issues, and simplifying the process to require only a single CAD file and an initial click from the user. 
By leveraging SAM2’s advanced segmentation capabilities and LightGlue’s feature point matching, our algorithm ensures stable and accurate 6D pose estimation across a wide range of objects. This has been validated through rigorous testing on the YCB dataset and industrial products, such as gears.

The system’s ability to maintain reliable tracking in dynamic environments, along with its fully automated operation after initial setup, makes it highly suitable for practical, real-world industrial applications.

\bibliography{GearTrack}
\bibliographystyle{icml2024}





\end{document}